\documentclass[twocolumn, DIV25, fontsize=9pt]{scrartcl}
\date{}
\pdfoutput=1

\usepackage{palatino}
\usepackage{graphicx}
\usepackage{subfigure}
\usepackage{float}
\usepackage{lettrine}
\usepackage{compactbib}
\usepackage{cite}
\usepackage{url}
\usepackage{float}
\usepackage{amsmath}
\usepackage[noend]{algpseudocode}
\usepackage{amsmath,amsfonts,amssymb}

\newif\iftodos

\todostrue

\usepackage[usenames,dvipsnames]{color} 
\iftodos

\newcommand{\todo}[1]{\textcolor{red}{[#1]}} 
\newcommand{\done}[1]{\textcolor{Emerald}{[#1]}} 
\newcommand{\comment}[1]{\textcolor{blue}{[#1]}} 
\newcommand{\jmccomment}[1]{\textcolor{magenta}{[#1]}} 
\newcommand{\lowpriority}[1]{\textcolor{green}{[#1]}} 
\newcommand{\todoOfficialVersion}[1]{} 
\else
\newcommand{\todo}[1]{} 
\newcommand{\done}[1]{} 
\newcommand{\comment}[1]{} 
\newcommand{\jmccomment}[1]{} 
\newcommand{\lowpriority}[1]{} 
\fi

\usepackage[labelfont={bf},small,sf]{caption}
\DeclareCaptionLabelSeparator{pp}{.}
\setcapindent{0cm}
\DeclareCaptionType{Supplementary figure}[Fig.]
\DeclareCaptionLabelFormat{bf-parens}{#1 S#2.}
\makeatletter
\renewcommand{\@biblabel}[1]{\quad#1.}
\makeatother

\RequirePackage{fancyhdr}
\title{\flushleft{\fontsize{21}{12}\selectfont \textsf{\textbf{Data-Efficient Exploration, Optimization, and Modeling of Diverse Designs through Surrogate-Assisted Illumination}}\\\smallskip
  \textbf{\large{\textsf{
      Adam Gaier$\mathsf{^{1}}$, Alexander Asteroth$\mathsf{^{1}}$, and Jean-Baptiste Mouret$\mathsf{^{2,3,4}}$}}}\\
  ~\\
         \small{
         $\mathsf{^1}$ Bonn-Rhein-Sieg University of Applied Sciences, Sankt Augustin, 53757, Germany\\
         $\mathsf{^2}$ Inria Nancy Grand - Est, Villers-l\'es-Nancy, F-54600, France\\
         $\mathsf{^3}$ CNRS, Loria, UMR 7503,  Vand\oe uvre-lÃšs-Nancy, F-54500, France\\
         $\mathsf{^4}$ Universit\'e de Lorraine, Loria, UMR 7503, Vand\oe uvre-l\'es-Nancy, F-54500, France\\
}}
\noindent{\normalsize \textsf{Preprint -- \today}}
\vspace*{-1.5cm}
}
\makeatletter
\def\@cite#1#2{$^{\mbox{\scriptsize #1\if@tempswa , #2\fi}}$}
\makeatother

\usepackage{algorithm,algorithmicx,algpseudocode}
\RequirePackage[normalem]{ulem} 
\RequirePackage{color}\definecolor{RED}{rgb}{1,0,0}\definecolor{BLUE}{rgb}{0,0,1} 

\begin{document}

\maketitle

\thispagestyle{fancy}
\pagestyle{fancy}

\begin{abstract}
The MAP-Elites algorithm produces a set of high-performing solutions that vary according to features defined by the user. This technique to 'illuminate' the problem space through the lens of chosen features has the potential to be a powerful tool for exploring design spaces, but is limited by the need for numerous evaluations. The Surrogate-Assisted Illumination (SAIL) algorithm, introduced here, integrates approximative models and intelligent sampling of the objective function to minimize the number of evaluations required by MAP-Elites. 

The ability of SAIL to efficiently produce both accurate models and diverse high-performing solutions is illustrated on a 2D airfoil design problem.
The search space is divided into bins, each holding a design with a different combination of features. In each bin SAIL produces a better performing solution than MAP-Elites, and requires several orders of magnitude fewer evaluations.
The CMA-ES algorithm was used to produce an optimal design in each bin:
with the same number of evaluations required by CMA-ES to find a near-optimal solution in a \emph{single} bin,
SAIL finds solutions of similar quality in \emph{every} bin.

\end{abstract}

\bigskip


\section*{Introduction}


\lettrine{C}{omputational} techniques for design optimization are often thought of by their creators as a final step in the design process. Imagining their techniques will be used to push the limits of performance, algorithm designers judge success by the ability to refine a design to its most optimal form~\cite{Hornby2006}.

If, however, the goal is truly to support designers, this sole emphasis on optimality may be misplaced. Autodesk~\cite{Bradner2014} recently conducted an interview to better understand how professional designers, engineers, and architects use design optimization tools. They found that optimization was most commonly used not at the end of the design process, but the beginning. Rather than using optimization to solve design problems, they were more commonly used to explore them. 

Generating a range of candidate solutions that represent different design alternatives allows designers to explore various design concepts, and examine the trade offs they represent. These generated designs provide insight into the assumptions and consequences inherent to the problem definition and constraints. Once constraints and objectives are reconsidered and adjusted, new designs are then generated and the process repeated.


This generative cycle allows designers to explore and describe complex design spaces, with high performing solutions acting as concrete way points. They can then manually iterate on the designs found through this collaborative human-computer exploration of the design space and, after consideration of intangibles such as aesthetics, finalize a design.

\begin{figure}[ht]
	\includegraphics[width=0.48\textwidth]{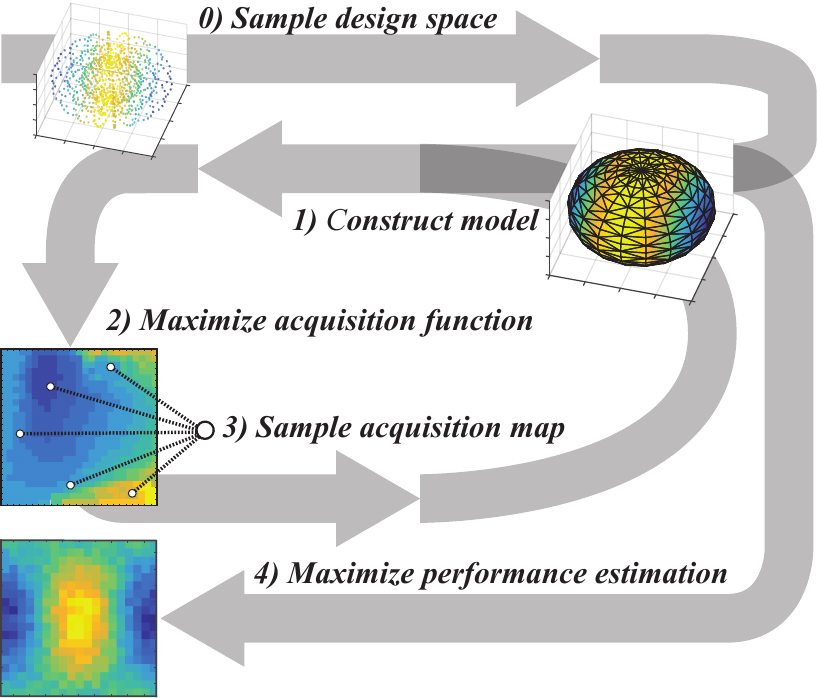}
	\centering
	\caption[caption]
	{
		\textit{Surrogate-Assisted Illumination (SAIL)} \\\hspace{\textwidth} 
		\textit{0)} Sample design space to produce initial solutions.
		\\\textit{1)} Construct model of objective function based on samples' performance.
		\\\textit{2)} Maximize the acquisition function, a balance of exploitation and exploration, in every region of the feature space, producing an \emph{acquisition map}.
		\\\textit{3)} Draw next samples to test on the objective from the \emph{acquisition map}. Repeat steps 1-3.
		\\\textit{4)} Maximize performance, as predicted by the resulting model, to produce a \emph{prediction map} populated with high performing designs in every region of the feature space.
	}
	\label{fig:alg_overview}
\end{figure}


Multi-objective optimization is perhaps the most commonly used tool to produce a variety of designs. When objectives are in conflict, each design in the Pareto front represents a trade-off between them~\cite{Deb2003}. 
However, during the explorative process interest for designers often lies not only in the maximization of objectives, but the effect of different design features on performance.

To probe the search space for interesting designs and design principles, new algorithms created specifically for design space exploration should be applied. One such algorithm, MAP-Elites~\cite{cully2015robots,Mouret2015} explicitly explores the relationship between user-defined features and performance. Designers select a few features deemed interesting or important, such as weight or structural strength, and MAP-Elites produces high-performing solutions which span the possible variations of those features. In this way this feature space is \emph{illuminated}, revealing the performance potential of each region of the feature space.

While effective at finding a variety of high-performing solutions, the number of evaluations required by MAP-Elites is immense.
The illumination process which produced the repertoire of hexapod controllers in~\cite{cully2015robots}, for example, required twenty million evaluations. In applications such as aerodynamic optimization, where a single evaluation can take hours, this is unrealistic.

In computationally expensive problems it is common to make use of surrogate models, approximate models of the objective function, that are based on previously evaluated solutions~\cite{Jin2005, Forrester2009, Shahriari2016}. 
These models are constructed through the sampling of solutions based on an \emph{acquisition function}, which balances exploitation and exploration to improve accuracy in high fitness regions.
These computationally efficient models can be used in place of the objective function during optimization, greatly accelerating the process. 
Incorporating surrogate-assistance into the evaluation-heavy illumination process has the potential to make MAP-Elites efficient enough for use in computationally expensive problems.


We present the Surrogate-Assisted Illumination (SAIL) algorithm to improve the efficiency, and so expand the applicability, of MAP-Elites. The value of integrating surrogate models into illumination relies on reducing computational cost while maintaining MAP-Elites' original capabilities, resulting in an algorithm that is:  

\begin{itemize}

	\item \textit{Divergent} - Produces a diversity of solutions which vary across a user-defined continuum;

	\item \textit{Accurate} - Predicts behavior of the objective function in high-performing regions;

	
	\item \textit{Optimal} - Produces high-performing solutions;
	
	\item \textit{Efficient} - Performs under computational constraints.

\end{itemize}

In broad terms SAIL works as follows (Figure~\ref{fig:alg_overview}, previous page): a surrogate model is constructed based on a set of initial
solutions and their measured performance. MAP-Elites is used
to produce solutions that maximize the acquisition function in
every region of feature space, producing an \emph{acquisition map}. New
samples are then drawn from the acquisition map and evaluated,
and these additional observations are used to improve the model.
This process is repeated to produce increasingly accurate models
of the high fitness regions of the feature space. Performance predictions
of the model can then be used by MAP-Elites in place
of the objective function to produce a \emph{prediction map} of estimated
optimal designs in every region of the feature space.

\section*{Related Work}
	\subsection*{Quality Diversity and MAP-Elites}\label{sec:mapElites}
Quality diversity (QD) algorithms~\cite{Pugh2016} use evolutionary methods to produce an archive of diverse, high quality solutions within a single run. 
Rather than seeking a single global optimum, QD algorithms discover as many types of solutions to a problem as possible, and produce a best possible example of each type.
For this reason they are also referred to as \emph{illumination} algorithms, as they illuminate the performance potential of different regions of the solution space.

Among the few illumination algorithms, novelty search with local competition (NSLC)~\cite{Lehman2011} uses a multiobjective approach to combine rewards for performance and novelty. 
The population is divided into niches based on similarity and their performance judged in relation to other members of their niche. Novelty is judged globally, with individuals rewarded based on their dissimilarity to their neighbors. In this way both exploration of the search space, as well as exploiting existing niches.

The MAP-Elites algorithm~\cite{cully2015robots,Mouret2015} is designed to produce high-performing solutions across a continuum of $n$ user-defined feature dimensions. It first divides the feature space into a grid, or map, of n-dimensional bins. The map houses the population of solutions, with each bin holding a single solution. When the map is visualized, with each bin colored according to the performance of the solution it contains, it provides an intuitive overview of the performance potential of each region of the feature space.

To initialize MAP-Elites a set of random solutions are first evaluated and assigned to bins. The bin location of a solution is based on its features. If, for example, the feature space is 2D with one dimension for weight and another for cost, a low cost and low weight solution would be placed in the low cost, low weight bin location of the map. If the bin is empty, the solution is placed inside. If another solution is already occupying the bin, the new solution replaces it if it has a higher fitness, otherwise it is discarded. As a result, each bin contains the best solution found so far for each combination of features. These solutions are known as \emph{elites}.

To produce new solutions, parents are chosen randomly from the elites, mutated, and then evaluated and assigned a bin based on their features. Child solutions have two ways of joining the breeding pool: discovering an unoccupied bin, or out-competing an existing solution for its bin. Repetition of this process produces an increasingly explored feature space and an increasingly optimal collection of solutions, \emph{illuminating} the performance potential of the entire feature space. MAP-Elites is summarized in Alg. \ref{alg:mapelites}.

\begin{algorithm} [h]
  \caption{MAP-Elites}
  \label{alg:mapelites}
  \begin{algorithmic}[1]
    \Function{MAP-Elites}{$objective\_function()$, $\mathcal{X}_{initial}$}
    \State $\mathcal{X} \gets \emptyset$  
    \Comment{empty map for genome}
    \State $\mathcal{P} \gets \emptyset$ 
    \Comment{empty map for performance}
   
    \State $\mathcal{X} \gets \mathcal{X}_{initial}$,
    $\mathcal{P} \gets objective\_function(\mathcal{X}_{initial})$ 

      \For{iter = $1 \to I$}
        \State $\mathbf{x~} \gets random\_selection(\mathcal{X})$
        \State $\mathbf{x'} \gets random\_variation(\mathbf{x})$
        \State $\mathbf{b'} \gets feature\_descriptor(\mathbf{x'})$
        \State $\mathbf{p'} \gets objective\_function(\mathbf{x'})$
        \If{$\mathcal{P}(\mathbf{b'}) = \emptyset$ or $\mathcal{P}(\mathbf{b'}) < \mathbf{p'}$}
          \State $\mathcal{P}(\mathbf{b'}) \gets \mathbf{p'}$, $\mathcal{X}(\mathbf{b'}) \gets \mathbf{x'}$
        \EndIf
      \EndFor
      \State \Return $(\mathcal{X}$, $\mathcal{P})$ 
      \Comment{Return illuminated map}
      \EndFunction
  \end{algorithmic}
\end{algorithm}

MAP-Elites has been shown to be effective in exploration and optimization in a variety of domains including: the design of walking soft robot morphologies~\cite{Mouret2015}, the generation of images that fool deep neural networks~\cite{nguyen2015deep}, and the evolution of robot controllers capable of adapting to damage~\cite{cully2015robots}. 

SAIL uses MAP-Elites rather than NSLC for illumination. 
While the niche definitions of NSLC are emergent, and neither even or consistent across runs, MAP-Elites defines a fixed structure of feature space boundaries, which greatly simplifies the process of sampling new solutions for inclusion in the surrogate model. 
Additionally, for design space exploration, this consistency allows designers to easily visualize and compare the effect of altered constraints and conditions on the feature space.

	\subsection*{Surrogate-Assisted Optimization}
Evolutionary approaches typically require a large number of evaluations before acceptable solutions are found. In many applications these performance calculations are far from trivial, and the computational cost becomes prohibitive. In these cases approximate models of the fitness function, or surrogates, are used in their place. 

Surrogate-assisted optimization has been a particularly useful approach in the computationally demanding context of computational fluid dynamics~\cite{Giannakoglou2006,Dumas2008}. In the context of MAP-Elites, even when evaluations are inexpensive, due to their sheer number surrogate-assistance has the potential to accelerate the illumination of the search space dramatically.

Modern surrogate-assisted optimization often takes place within the framework of Bayesian optimization (BO)~\cite{Brochu2010, cully2015robots, Shahriari2016}. BO approaches the problem of optimization not just as finding the most optimal solution, but of modeling the underlying objective function in high-performing regions. 
To estimate the objective function probabilistic models are used, giving each sample a predicted objective value and a certainty in that prediction. New samples are chosen where the model predicts a high objective value (exploitation) and where prediction uncertainty is high (exploration). The relative emphasis on exploitation and exploration is determined by the \emph{acquisition function}. The sample which maximizes the acquisition function is chosen as the next observation.

A variety of data-driven machine learning techniques such as polynomial regression, support vector machines, and artificial neural networks, can be used to construct surrogate models~\cite{Forrester2009,Jin2005}, however as BO requires a probabilistic model, Gaussian processes (GP)~\cite{Rasmussen2006} are typically used.

	\subsubsection*{Gaussian Process Models}\label{sec:gp}

In the presented implementation of SAIL, Gaussian process (GP) models~\cite{Rasmussen2006} are chosen for fitness approximation. GP models are effective even with a small number of samples and their predictions include a measure of certainty. In the active learning context of surrogate-assisted optimization a measure of model uncertainty is particularly useful, as this allows for the balancing of exploration and exploitation.

Gaussian process models are a generalization of the Gaussian distribution: where a Gaussian distribution describes random variables, defined by mean and variance, a Gaussian process describes a random distribution of functions, defined by a mean function $m$, and covariance function $k$. 

\begin{equation}
f(x) \sim GP(m(x), k(x,x'))
\end{equation}

In much the same way as an artificial neural network can be thought of as a function that returns a scalar given an arbitrary input vector $x$, a GP model can be thought of as a function that, given $x$ returns the mean and variance of a normal distribution, with the variance indicating the certainty of our prediction.

Gaussian process models make their predictions based on locality in the input space, a relationship defined by a covariance function. A common choice is the squared exponential function:  the closer the points are in input space the more closely correlated they are in the output space:

\begin{equation}
k(\mathbf{x_i,x_j}) = \exp{ \Big( -\frac{1}{2} \|\mathbf{x}_i - \mathbf{x}_j\|^2 \Big)}
\end{equation}
Given observations $D = ({x_{1:t}, f_{1:t}})$ where $f_{1:t} = f(x_{1:t})$,  we can build a matrix of covariances. In the simple noise-free case we can then construct the kernel matrix:
\begin{equation}
K = 
 \begin{bmatrix}
  k(x_1,x_1) 	& \cdots  	& k(x_1,x_t) 	\\
  \vdots  		& \ddots  	& \vdots 		\\
  k(x_t,x_1) 	& \cdots  	&  k(x_t,x_t)
 \end{bmatrix}
\end{equation}

Considering a new point ($x_{t+1}$) we can derive the value ($f_{t+1} = f(x_{t+1})$) from the normal distribution (for simplicity we assume a zero mean function $m(x)= 0$):

\begin{equation}
 \begin{bmatrix}
  \mathbf{f}_{1:t}\\
  f_{t+1}
 \end{bmatrix}
  \sim \mathcal{N} \bigg( \mathbf{0},
 \begin{bmatrix}
  K 	& \mathbf{k} 	\\
  \mathbf{k}^T 	& k(\mathbf{x}_{t+1}, \mathbf{x}_{t+1})
 \end{bmatrix}
 \bigg)
\end{equation}
where $\mathbf{k} = [
  k(\mathbf{x}_{t+1},\ \mathbf{x}_1), 
  k(\mathbf{x}_{t+1},\  \mathbf{x}_2), 
  \ \ldots\ , 
  k(\mathbf{x}_{t+1},\  \mathbf{x}_t)]^T$
allowing us to compute the GP as:
\begin{equation}
P(f_{t+1}|D_{1:t},x_{t+1}) = 
\mathcal{N} \Big( \mu_t (x_{t+1}), \sigma_t^2 (x_{t+1}) \Big)
\end{equation}
where: 
\begin{align}
\mu_t(x_{t+1}) &= \mathbf{k}^T \mathbf{K}^{-1} \mathbf{f}_{1:t} \\
\sigma_t^2(x_{t+1}) &= k(\mathbf{x}_{t+1}, \mathbf{x}_{t+1}) - \mathbf{k}^T \mathbf{K}^{-1}\mathbf{k}
\end{align}
gives us the predicted mean and variance for a normal distribution at the new point $x_{t+1}$. If we were then to evaluate the objective function at this point, we would add it to our set of observations $D$, reducing the variance at $x_{t+1}$ and at other points near to $x_{t+1}$.

In this pure generalized form, our GP model weighs variations in every dimension equally, applying the same squared exponential relationship regardless of input dimension. For higher dimensional problems each dimension's effect on the output is also weighted via a technique known as automatic relevance detection (ARD). The hyperparameters which weigh each dimension are set by maximizing the likelihood of the model given the data~\cite{Rasmussen2006}. This increases model accuracy, and the weighting provides an understandable estimation of the relative importance of each input dimension.

\section*{Surrogate-Assisted Illumination}
To understand the relationship between features and performance, SAIL models the underlying objective function in different regions of the feature space.
Sampling of the objective function in order to model its behavior in the best performing regions is also the goal of Bayesian optimization~\cite{Brochu2010, Shahriari2016}, and we adopt similar methods.

Bayesian optimization has two components. The first is a probabilistic surrogate model of the objective function, which in SAIL takes the form of a Gaussian process (GP) model (see Section \ref{sec:gp}). The second is an acquisition function, which describes the utility of sampling a given point. The point with maximal utility is evaluated and its performance added to an observation set. The updated set of observations is then used to produce a more informed GP model. As we are not looking to model the objective function only at the global optimum, but at optima in all regions of the feature space, we must produce points which maximize utility in every region of the feature space.


Evaluating new solutions is expensive, making the definition of ``utility'' critical to performance. Balance must be maintained between exploration, sampling points with high uncertainty, and exploitation, sampling of points which are likely to perform better than our current solutions.

The acquisition function defines how the balance between exploration and exploitation is determined. In SAIL, the \textit{upper confidence bound} (UCB)~\cite{Srinivas2009} is used. Proposed as part of the GP-UCB algorithm, use of UCB has been shown to minimize regret and maximize information gain in multi-armed bandit problems~\cite{Srinivas2009}. UCB judges potential observations optimistically, favoring uncertainty under the assumption that higher uncertainty hides a potentially higher reward. A high mean~($\mu(x)$) and large uncertainty ($\sigma(x)$) are both favored, with relative emphasis tuned by the parameter $\kappa$. 
\begin{equation} \label{eq:UCB}
  UCB(\mathbf{x}) = \mu(\mathbf{x}) + \kappa\sigma(\mathbf{x})
\end{equation}

UCB performs competitively with more complex acquisition functions such as Expected Improvement (EI) and Probability of Improvement (PI) \cite{Brochu2010,Calandra2013}. These acquisition functions rely on comparisons to the current optimum, while UCB is based only on the underlying model. As SAIL is used to solve numerous localized problems in parallel, it requires an acquisition function independent of the global optimum. 
If compared globally, solutions in less optimal regions of the design space would have a vanishingly small probability of improving on the global optimum, and as bins are likely not to contain any precisely evaluated solutions, it will not always be possible to perform local comparisons against optima within a bin.


To estimate the relationship between features and performance, SAIL models the objective function not only around a global optimum, but around high-performing solutions over the entire feature space. 
To accurately predict performance in this slice of the search space, we produce potential observations with every combination of features.
By dividing the feature space into bins and using MAP-Elites to produce a solution which maximizes the acquisition function in each, we produce an \emph{acquisition map}. 

It is from the acquisition map that we draw new observations.
To reduce uncertainty over the entire feature space we use a Sobol sequence~\cite{Niederreiter1988} to select which bins to draw the next samples from. 
Sobol sequences iteratively divide the range into finer uniform partitions, allowing for even sampling across the feature space.
In the case that a sampled point results in an invalid solution, the next in the sequence can be drawn.
Once evaluated the performance of these samples can be added to our set of observations and a new GP model constructed. A new acquisition map can then be created using this updated model, and the process repeated.

\begin{algorithm} [h]
  \caption{Surrogate-Assisted Illumination}
  \label{alg:sail}

  \begin{algorithmic}[1]
    \\\Comment{\textit{Initialize with G solutions drawn from Sobol sequence}}
    \State $\mathcal{X} \gets Sobol_{1:G}$,
    $\mathcal{P} \gets PE(\mathcal{X} )$ 
    \Comment{\textit{PE = precise evaluation}}
    
    \\\State{\textbf{1) Produce Acquisition Map}}
    \For {iter $ = 1 \to precise\_evaluation\_budget $}
    
    \State $\mathcal{D} \gets (\mathcal{X}, \mathcal{P})$ 
    \Comment{Observation Set: \textit{Genome, Performance}}

    \State $\mathcal{GP} \gets Gaussian\_process\_model(\mathcal{D})$
    \State $acquisition() \gets UCB(\mathcal{GP}(x))$
    \State $(\mathcal{X}_{acq}, \mathcal{P}_{acq}) =$ 
    \Call{MAP-Elites}{$acquisition(), \mathcal{X}$}

    \Comment{\textit{Select solutions from acquisition map for PE}}
    \State $\mathbf{x} \gets \mathcal{X}_{acq}(Sobol_{iter})$
    \State $\mathcal{X} \gets \mathcal{X} \cup \mathbf{x}$,
    $\mathcal{P} \gets \mathcal{P} \cup PE(\mathbf{x})$
    \EndFor
    \\\State{\textbf{2) Produce Prediction Map}}
    \State $\mathcal{D} \gets (\mathcal{X}, \mathcal{P})$ 
    \Comment{Observation Set: \textit{Genome, Performance}}

    \State $\mathcal{GP} \gets Gaussian\_process\_model(\mathcal{D})$
    \State $prediction() \gets mean(\mathcal{GP}(x))$

    \State $(\mathcal{X}_{pred}, \mathcal{P}_{pred}) =$ 
    \Call{MAP-Elites}{$prediction(), \mathcal{X}$}

  \end{algorithmic}
\end{algorithm}

The SAIL algorithm is more precisely defined in (Alg.\ref{alg:sail}). An initial set of individuals is created using a Sobol sequence~\cite{Niederreiter1988} to ensure initially even coverage of the \emph{parameter} space. These individuals and their performance form a set of observations $D$, which is used to construct a GP model. An empty acquisition map is then created and filled with the individuals from $D$, along with their utility as judged by the acquisition function. These individuals are taken as the starting population for MAP-Elites (Alg.\ref{alg:mapelites}) which then illuminates the map as described in Section~\ref{sec:mapElites}: an elite is selected and mutated to produce a child, it is assigned a bin based on its features, and it finally competes for the bin if it is not occupied. This illumination process repeats for a number of iterations, and results in an acquisition map of elite individuals who maximize the acquisition function in their bin.

From the acquisition map we select the next samples for evaluation. 
To ensure even coverage of the \emph{feature} space, we again employ a Sobol sequence to direct the sampling, this time producing coordinates in feature space rather than parameter values. These coordinates indicate the bin to be sampled, and the individual stored is precisely evaluated.
Once evaluated these new individuals and fitness pairs are added to our observation set $D$ and the process can be repeated.

The mean prediction of the resulting GP model can then be taken as the fitness function of MAP-Elites, and a \emph{prediction map} produced. This map is an estimate of the relationship between features and performance, including an optimal design for each bin. As only the surrogate model of the objective is used for evaluation, this prediction map can be produced with minimal computation.

 \label{sec:sail}

\section*{Experimental Setup}
	\subsection*{Objectives and Constraints}\label{sec:objectives}

We evaluate the performance of SAIL on a classic design problem, 2D airfoil optimization. Fitness is defined as minimal drag while maintaining the same area and not decreasing lift compared to a base airfoil. Quadratically increasing penalties are introduced into the fitness function to ensure that these constraints are followed with little deviation. The high-performing RAE2822 airfoil was chosen as our base, with foils evaluated at an angle of attack of $2.7^{\circ}$, at Mach 0.5 and Reynolds number of $10^6$. Evaluation criteria are formally defined for a solution $x$ as:

\begin{equation}
\mathit{fitness}(x) = \mathit{drag}(x) \times \mathit{penalty}_{\mathit{lift}}(x) \times \mathit{penalty}_{\mathit{area}}(x)
\end{equation}
where $\mathit{drag}(x) = -log(C_D(x))$

\begin{equation}
    \mathit{penalty}_{\mathit{lift}}(x)= 
	\begin{cases}
	    \Big(\frac{C_L(x)}{\mathit{lift}_{\mathit{base}}}\Big)^2,& \text{if } C_L(x) < lift_{base}\\
	    1,              & \text{otherwise}
	\end{cases}
\end{equation}

\begin{equation}
    \mathit{penalty}_{\mathit{area}}(x)= \Big(1-\frac{|\mathit{area} - \mathit{area}_{\mathit{base}}|}{\mathit{area}_{\mathit{base}}}\Big)^7
\end{equation}

While the area of the foil can be directly measured without aerodynamic tests, the drag\footnote{As $C_D$ values are very small, they are converted to log scale in our fitness calculation} ($C_D$) and lift ($C_L$) must both be approximated. The UCB of the drag prediction is taken as the drag component of our fitness function:
\begin{equation}
drag(x)' = \mu_{\text{drag}}(x) + \kappa\sigma_{\text{drag}}(x)
\end{equation}
As individuals are not rewarded for having high lift, but are only expected to maintain performance, we treat the prediction problem as one of classification rather than regression. Individuals are penalized based on the probability that they will have a lower lift than our base foil, based on the mean and variance supplied by our GP model:
\begin{equation}
\mathit{penalty}_{\mathit{lift}}(x)'= 1-P(C_L(x) < lift_{base})
\end{equation}

\subsection*{Representation}

We encode the airfoil using a variation of the the airfoil-specific PARSEC parameterization~\cite{Sobieczky1999}. PARSEC uses polynomial expressions to encode design features, such as the radius of the leading edge or the curvature of the upper surface, requiring a small number of design parameters to express a large variety of designs.

\begin{figure}[h]
	\centering
	\includegraphics{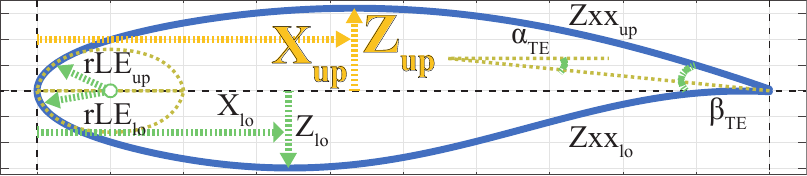}
	\caption
	{
		The ten parameters used to define an airfoil. Dimensions of variation ($X_{up}$ and $Z_{up}$) in gold.
	}
	\label{fig:parsec}
\end{figure}

We restrict the design space to foils with trailing edges which have the same end point and sharpness as our base foil. We also add an additional degree of freedom by splitting the leading edge radius into an upper ($r_{LE_{up}}$) and lower leading edge radius ($r_{LE_{lo}}$). The ten parameters used to define an airfoil are shown in Figure~\ref{fig:parsec}.

	\subsection*{Dimensions of Variation} \label{sec:varDims}

Illumination algorithms allow us to define dimensions of variation in which we would like to explore. We choose two of our PARSEC descriptors: the height of the highest point on the top side of the foil ($Z_{up}$), and the location along the length of the wing of this highest point ($X_{up}$). In early tests these parameters were found to be highly predictive of the drag. The range of $Z_{up}$ and $X_{up}$ are discretized into 25 partitions, giving us a 25$\times$25 grid, or 625 bins.

In practice the dimensions of variation do not have to be parameter values, and in fact it is desirable that they not be. Defining dimensions of variation which do not align with the representation, but rather correspond to more abstract feature measures, allows for search in a low-dimensional feature space even with a high dimensional representation. Low level features should be chosen based on characteristics that the user would like to explore or, through their own experience, know are important or interesting. In this case parameter values were chosen as dimensions of variation for ease of analysis and comparison with other algorithms.

	\subsection*{Baseline and Hyperparameters} \label{sec:baseline}

To evaluate the optimality of the prediction maps produced by SAIL and how efficiently they are produced we compare to 1) standard MAP-Elites without surrogate assistance, and two variants of a traditional convergent search algorithm: 2) the covariance matrix adaptation evolution strategy (CMA-ES), and 3) surrogate-assisted CMA-ES (SA-CMA-ES). The unit of comparison used is the number of precise evaluations (PE), i.e. actual calls to the simulator.

We provide the SAIL algorithm a computational budget of 1000PE. 50PE is used to evaluate the initial pool of individuals which form the basis of the GP model. The remaining 950PE are spent in the course of the algorithm, with 10 new individuals added to the observation set at every iteration (Alg. \ref{alg:sail} lines 4-12). This was compared to the standard MAP-Elites algorithm with a budget of $10^5$PE.

We are unaware of any other similar design space exploration techniques and so for a better understanding of the difficulty of the task and the optimality of the solutions produced by SAIL we compare to the results of traditional convergent search algorithms, algorithms which are designed to find a single optimum solution. 
As we have chosen parameter values as our dimensions of variation, it is possible to confine a search within one bin of the map by restricting the valid parameter ranges of $X_{up}$ and $Z_{up}$.
Each bin can then be thought of as a single search problem. We solve each of these subproblems with the well-established covariance matrix adaptation evolution strategy (CMA-ES)~\cite{Hansen2001}. A budget of 1000PE \emph{per bin} is given to find optimal solutions. 

A surrogate-assisted variant of CMA-ES (SA-CMA-ES) is also applied to solve the subproblem in each bin. 
A GP model is produced with 25 initial individuals drawn from a Sobol sequence, sampling in the same way as SAIL. CMA-ES is then used to maximize the acquisition function, computed with the same UCB-based fitness criteria as SAIL, described in Section \ref{sec:objectives}. The found optimum is added to the set of observations and the optimization process repeats with an updated model. This process is repeated 75 times, for a total of 100PE. Each bin is considered a distinct subproblem, and models and samples are not shared across bins.

Runs of CMA-ES, SA-CMA-ES, SAIL, and MAP-Elites were each replicated 20 times.\footnote{One replicate, including data gathering, with 8 cores of a Intel Xeon 2.6GHz processor required:
SA-CMA-ES:~32h, CMA-ES:~80h, SAIL:~12h, MAP-Elites:~14h}
As optimal performance varies depending on the bin, in some comparisons fitness will be reported as a percentage of the optimum value found in all experiments, i.e. $0\%$ - $100\%$ of the optimum.
Unless otherwise mentioned all values are medians across all experiments.
Valid initial designs with a highest point at the leading edge of the wing (high $Z_{up}$ and low $X_{up}$) could not be found due to geometric constraints inherent in the PARSEC representation~\cite{Padulo2009}. Only the remaining $577$ bins were considered.
Beyond our own implementation\footnote{\url{github.com/agaier/sail_gecco2017}} standard implementations were used for CMA-ES~\cite{cmaes}, Gaussian Processes~\cite{gpml}, and airfoil simulation~\cite{xfoil}.

\section*{Results}
\begin{figure}[h]
	\centering
	\includegraphics{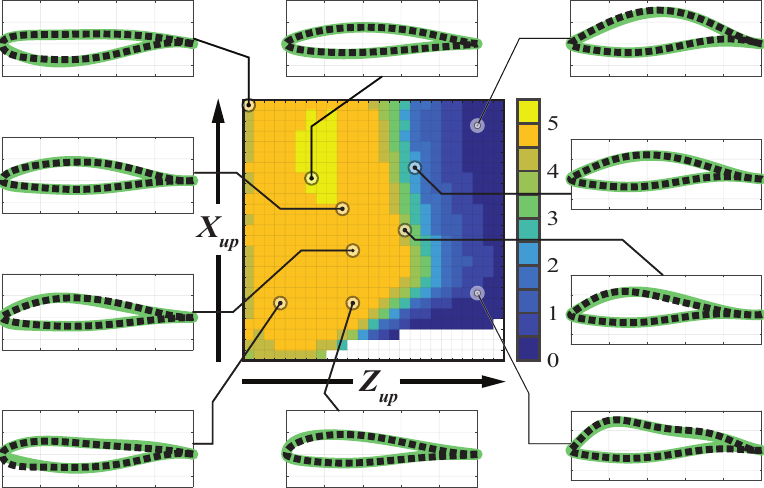}
	\caption
	{
		\textit{Design Space Overview with SAIL}\newline
		Prediction map produced by SAIL after 1000PE.\\ Border: Median performing designs found by SAIL in green, best designs found by CMA-ES in black.
	}
	\label{fig:designMap}
\end{figure}


The prediction map of the feature space produced by SAIL in Figure~\ref{fig:designMap} visualizes the effect of the explored features ($X_{up}$ and $Z_{up}$) on performance.
The height of the airfoil ($Z_{up}$) has the strongest effect on fitness, with taller airfoils performing worse than flatter airfoils. The location on the wing of the highest point ($X_{up}$) has a more nuanced effect, increasing or decreasing fitness depending on the height of the airfoil.
The best performing foils are not at the extremes of the feature space, but at a peak within the mid ranges. Similar designs and trends were also found by CMA-ES.

	\subsection*{Accuracy}
\begin{figure}[ht]
	\centering
	\includegraphics{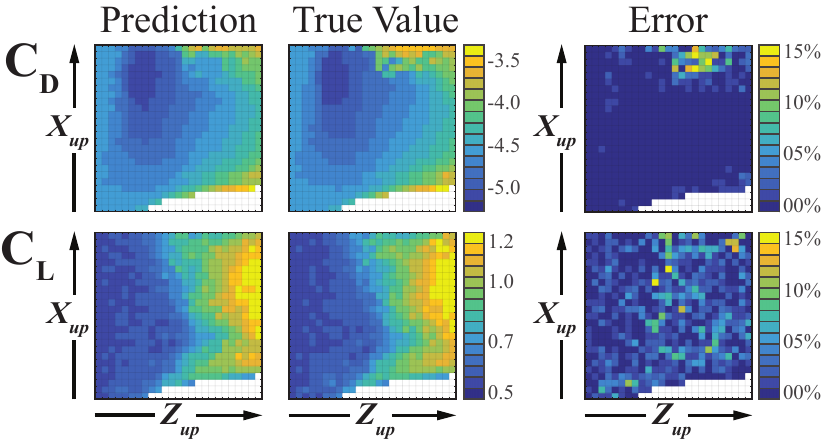}
	\caption
	{
		\textit{Drag and Lift Predictions Per Bin}\newline
		Predicted and true values of drag ($log(C_D)$) and lift ($C_L$) for designs in each bin after 1000PE.
	}
	\label{fig:acc_maps}
\end{figure}

To evaluate the accuracy of the produced models, after the final sample was collected a prediction map was produced. Each design in the prediction map was then precisely evaluated and the true $C_D$ and $C_L$ compared to the prediction of the model. The median results are shown in Figure~\ref{fig:acc_maps}. On the majority of samples the surrogate is reliably accurate, with more than 90\% of drag ($log(C_D)$) predictions and more than 80\% of lift ($C_L$) predictions within 5\% of their true value. Drag errors are clustered in the same region of design space, a region where the flow simulator was less likely to converge and produce valid results.

\begin{figure}[ht]
\centering
	\includegraphics{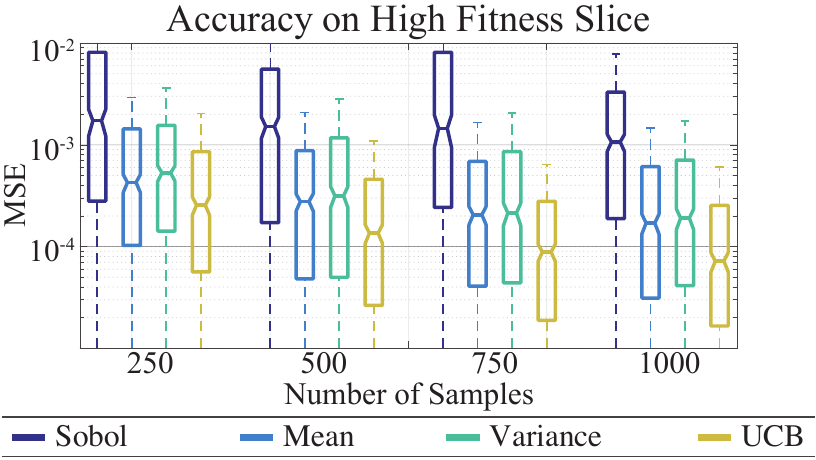}
	\caption
	{
		\textit{Accuracy of Sampling Strategies}\newline
		Mean squared error (log scale) of drag prediction on optimal designs. Models constructed using designs sampled from parameter space using a Sobol sequence or selected from acquisition maps produced with the mean, variance, or the UCB of the prediction. 
	}
	\label{fig:modelComparison} 
\end{figure}

The purpose of our models is to estimate performance in the optimal regions of the search space. 
To test their accuracy in this high fitness slice, we measure their ability to predict the performance of the best designs found by CMA-ES in each bin.
We compare models built using a naive sampling of the parameter space with a Sobol sequence~\cite{Niederreiter1988} to sampling done using acquisition maps produced by SAIL. These acquisition maps are produced by maximizing three different acquisition functions: the mean or variance alone, and the UCB, a combination of the mean and variance (see Section:~\ref{sec:sail}).
%
The accuracy of each model's drag prediction on the best design in every bin is then measured at various stages of the sampling process~(Figure~\ref{fig:modelComparison}).

By concentrating the sampling process on either high-performing solutions or on reducing overall uncertainty we are able to produce better performing models than evenly sampling the parameter space. When both uncertainty and performance are considered when using the UCB, SAIL produces models that are an order of magnitude more accurate than uniform sampling.

	\subsection*{Optimality and Efficiency}


Though our goal is not to directly compete with algorithms designed to find one optimal solution, to accurately portray the design space it is critical that the solutions found are representative of the best designs in their region. 

\begin{figure}[ht]
	\centering
	\includegraphics{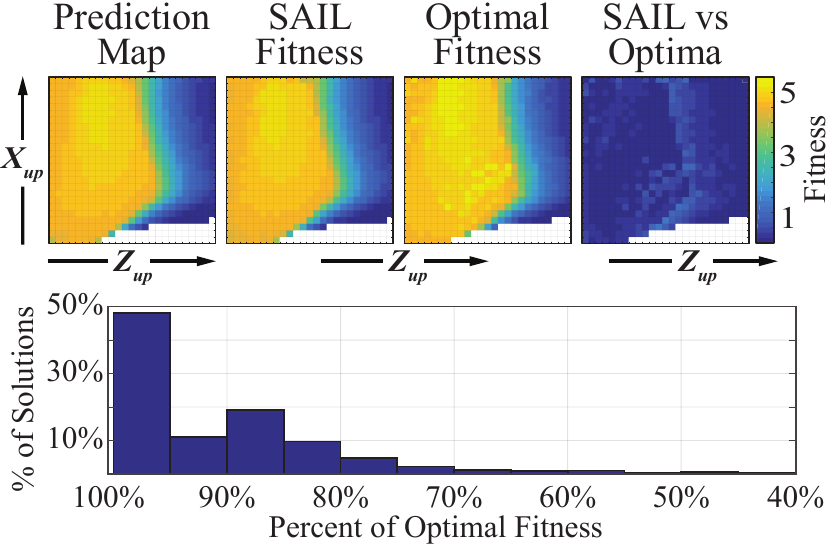}
	\caption
	{
		\textit{Performance of Designs Found By SAIL}\newline
		\textit{Top}: Median predicted and true performance of designs found by SAIL with a budget of 1000PE compared with performance of optimal designs found by 20 runs of CMA-ES per bin ($\approx$11.5~$\mathbf{million}$ PE) 
		\newline
		\textit{Bottom}: Optimality of SAIL designs per bin.
	}
	\label{fig:opt_maps}
\end{figure}

We compared the designs found in each bin by SAIL after 1000PE to the best design found by CMA-ES after all 20 runs for 1000PE in each valid bin ($\approx11.5~$million PE in total). Figure~\ref{fig:opt_maps} shows the median values of the prediction map, the true performance of those median designs, the optimal performance found after 20 runs of CMA-ES, and the fitness difference between these optimal values and those found by SAIL. The fitness potential of the feature space is well illuminated: found designs perform within 5\% of the optimum in nearly half of bins, and the relationship between features and performance is accurately portrayed.

\begin{figure*}
	\centering
	\includegraphics[width=1\textwidth]{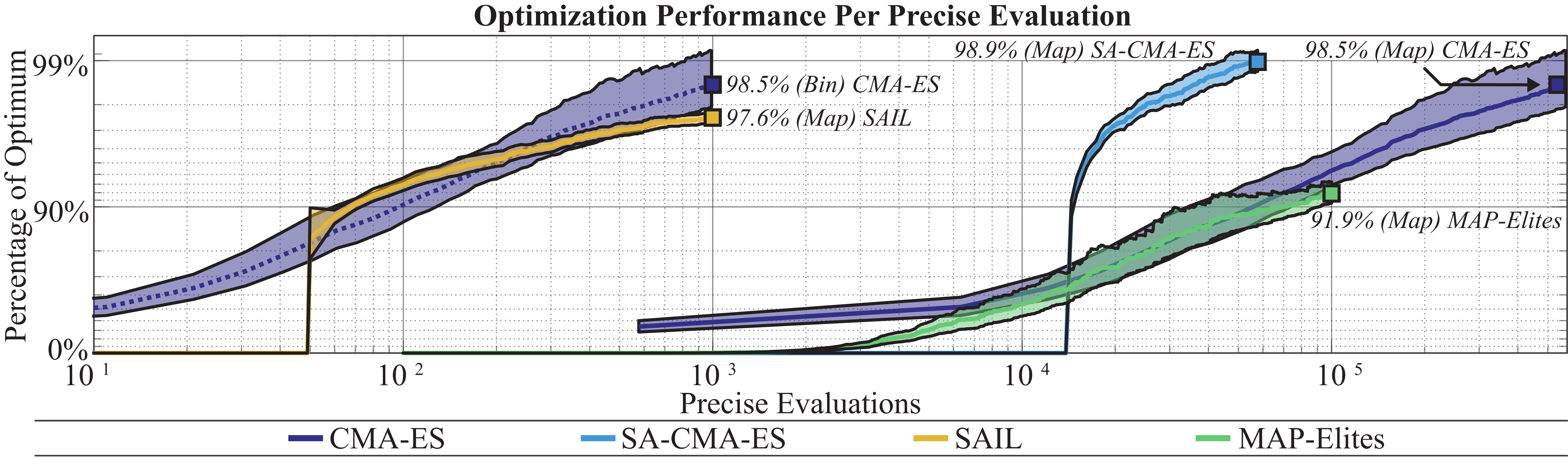}
	\caption
	{
		\textit{Optimization Efficiency in a Single Bin and Over the Entire Design Space}\newline
		Computational efficiency of CMA-ES, SA-CMA-ES, MAP-Elites, and SAIL in precise evaluations. \emph{Bin}: median progress towards optimum in every bin. \emph{Map}: performance of CMA-ES and SA-CMA-ES is median bin performance multiplied by number of bins. 
		Performance of individuals produced to construct initial models is set to 0\%.
		Bounds indicate one standard deviation over 20 replicates. Precise evaluations and performance in log scale.
	}
	\label{fig:progress}
\end{figure*}

As we have found no similiar design space exploration algorithms beyond MAP-Elites for comparison, to judge the efficiency of our algorithm we turn to convergent search techniques. As CMA-ES was not intended for use across a multitude of subproblems the total number of PEs needed to arrive at an optimized feature map is highly dependent on the number of bins in the map. 
Therefore we also compare SAIL to the performance of CMA-ES in a single bin. The progress of the different approaches is compared in Figure \ref{fig:progress}. 

Single bin performance is taken as the median performance over all bins. Optimization may progress faster or slower depending on the bin, and this gives us a measure of how near an average bin will be to the optimum after a given number of precise evaluations. Map performance is simply this median multiplied by the number of bins. Performance of individuals produced by SA-CMA-ES and SAIL to construct the initial models is set to 0\%, with the first valid performance indicators at 25PE/bin and 50PE (total) respectively. 

With the same computational budget required by CMA-ES to find a near optimal solution in a \textit{single} bin, SAIL produces solutions of similar quality in \textit{every} bin.

The acceleration afforded by surrogate modeling has an even more pronounced effect on the divergent search techniques (MAP-Elites and SAIL) than on the convergent approaches. Incorporating surrogate-assistance into CMA-ES improves performance by an order of magnitude. MAP-Elites, even when given two orders of magnitude more precise evaluations, is still unable to compete with SAIL's performance. Surrogate-assisted optimization allows for estimations of performance to be calculated based on similarity of solutions, a technique which fits neatly into the illumination approach as solutions in close proximity on the map are also likely to perform similarly.

\section*{Conclusion and Discussion}
	The SAIL algorithm produces a model of the objective function in high-performing regions across the feature space despite a limited computational budget.
	With the knowledge that our models are accurate, we can be confident in the prediction map's depiction of the feature space, even if the solutions in the map have not been precisely evaluated.

	Prediction maps which illuminate different feature combinations of the search space can be produced quickly without additional evaluations or model training.
	This allows easy exploration and visualization of the design space through various lenses.
	Acceleration of the illumination process allows the exploration process to take place in an anytime fashion: as soon as new samples are evaluated, the surrogate model can be reconstructed and estimates of the entirety of the feature space can be rapidly updated.

	This assumes, of course, that our models can be trained quickly. In our analysis we concentrated only on the efficiency of the algorithm with regards to precise evaluations. While appropriate in extreme cases, such as fluid dynamics, in practice the cost of training surrogate models must be balanced against the savings they yield. In light of the sheer number of evaluations required by MAP-Elites the savings will typically be substantial.


	While directing the sampling process with the UCB of the prediction produced more accurate models than using the mean or variance alone, the importance of this improved accuracy is unclear. More investigation is needed into the effect of different acquisition functions and how best to then choose samples from the resulting acquisition map. In the most expensive cases, human-in-the-loop approaches may be appropriate, with experienced designers selecting designs from the acquisition map for evaluation.

	In our experiments parameter values served as features, making search within regions of the feature space with a traditional optimizer straightforward. Features are not always so easy to compute, especially if those features are behaviors identified during evaluation, as in evolutionary robotics~\cite{Mouret2012}. In cases where classifying a solution in feature space is itself expensive, it may be necessary to also construct models to approximate the features of a new individual. 

	MAP-Elites grew out of the evolutionary robotics community where it is common to employ representations that themselves evolve and grow more complex, such as NEAT~\cite{Stanley2002}. If SAIL is to be used with non-static representations, like those produced by NEAT, or those that are static but very high dimensional, like those produced by CPPNs~\cite{Stanley2009}, specialized surrogate modeling techniques must be developed.

	Though MAP-Elites has shown remarkable potential, the intensive computation it requires precludes its use in many domains. 
	By pairing MAP-Elites with a surrogate modeling, a Bayesian optimization equivalent for illumination is created. By enabling illumination in computationally expensive domains 
	SAIL opens up new avenues for experiments and applications of quality-diversity techniques.

\subsection*{Source code} The source code used to produce the results in this publication is available with an open-source license on Github at: \url{http://www.github.com/agaier/sail_gecco2017}

\begin{small}
  \sffamily
  \bibliography{sail}
\end{small}

\section*{Acknowledgments}
This work received funding from the European Research Council (ERC) under the European Union's Horizon 2020 research and innovation programme (grant agreement number 637972, project "ResiBots") and the German Federal Ministry of Education and Research (BMBF) under the Forschung an Fachhochshulen mit Unternehmen programme  (grant agreement number 03FH012PX5 project "Aeromat"). The authors would like to thank Roby Velez, Alexander Hagg, and the ResiBots team for their feedback on this paper.

\end{document}